\documentclass[12pt, preprint,authoryear]{elsarticle}
\usepackage{graphicx}
\usepackage{amsmath}
\usepackage{amsfonts}
\usepackage{amssymb}
\usepackage{amsthm} 
\usepackage[outline]{contour} 
\usepackage{booktabs}
\usepackage{multirow}
\usepackage{geometry}
\usepackage[flushleft]{threeparttable}

\newtheorem{definition}{Definition}[section]
\newtheorem*{remark}{Remark}
\journal{Neural Networks}

\begin{document}

\begin{frontmatter}
\title{Separable Gaussian Neural Networks: Structure, Analysis, and Function Approximations}
\author[CPSU]{Siyuan Xing}
\affiliation[CPSU]{organization={Department of Mechanical Engineering, California Polytechnic State University},
city={San Luis Obispo},
state={California},
postcode={93407},
country={USA}}
\author[UCM]{Jian-Qiao Sun}
\affiliation[UCM]{organization={Department of Mechanical Engineering, School of Engineering, University of California},
city={Merced},
state={California},
postcode={95343},
country={USA }}
\cortext[UCM]{Corresponding author: jqsun@ucmerced.edu}

\begin{abstract}
The Gaussian-radial-basis function neural network (GRBFNN) has been a popular choice for interpolation and classification. However, it is computationally intensive when the dimension of the input vector is high. To address this issue, we propose a new feedforward network - Separable Gaussian Neural Network (SGNN) by taking advantages of the separable property of Gaussian functions, which splits input data into multiple columns and sequentially feeds them into parallel layers formed by uni-variate Gaussian functions. This structure reduces the number of neurons from $O(N^d)$ of GRBFNN to $O(dN)$, which exponentially improves the computational speed of SGNN and makes it scale linearly as the input dimension increases. In addition, SGNN can preserve the dominant subspace of the Hessian matrix of GRBFNN in gradient descent training, leading to a similar level of accuracy to GRBFNN. It is experimentally demonstrated that SGNN can achieve 100 times speedup with a similar level of accuracy over GRBFNN on tri-variate function approximations. The SGNN also has better trainability and is more tuning-friendly than DNNs with RuLU and Sigmoid functions. For approximating functions with complex geometry, SGNN can lead to three orders of magnitude more accurate results than a RuLU-DNN with twice the number of layers and the number of neurons per layer. 
\end{abstract}

\begin{keyword}
Function approximations \sep Separable Gaussian Neural Networks \sep
\sep Gaussian-radial-basis functions \sep Separable functions \sep Subspace gradient descent
%
\end{keyword}


%

\end{frontmatter}

\section{Introduction}

Radial-basis functions have many important applications in the fields such as function interpolation \citep{Dynetal1986}, meshless methods \citep{Duanmeshless2008}, clustering classification \citep{WuK-mean2012}, surrogate models \citep{AkhtarShoemaker2016}, Autoencoder \citep{Daoud2019}, and dynamic system design \citep{YuRBFNDSD2011}. The Gaussian-radial-basis-function neural network (GRBFNN) is a neural network with one hidden layer and produces output in the form
\begin{equation}
\tilde{f}(\mathbf{x})=\sum_{k=1}^{N} W_kG_k(\mathbf{x}),
\label{eq:GRBFNN}
\end{equation}
where $G_k(\mathbf{x})$ is a radially-symmetric unit represented by the Gaussian function such as
\begin{equation}
G_k(\mathbf{x})=\exp\left(-\frac{1}{2\sigma_k^2}||\mathbf{x}-\boldsymbol{\mu}_k||^2\right).
\label{eq:sec=1:gaussian_radial_basis_units}
\end{equation} 
Herein, $\boldsymbol{\mu}_k$ and $\sigma_k$ are the center and width of the unit that can be tuned to adjust its localized response. The locality is then utilized to approximate the output of a nonlinear mapping through the linear combination of Gaussian units. Although it has been shown that GRBFNN outperforms multilayer perceptions (MLP) in generalization \citep{TaoKMRBF1993}, tolerance to input noise \citep{moody_fast_1989}, and learning efficiency with a small set of data \citep{moody_fast_1989}, the network is not scalable for problems with high-dimensional input, because the neurons in need for accurate predictions and the corresponding computations exponentially increase with the rise of dimensions. This paper aims to tackle this issue and make the network available for high-dimensional problems.

GRBFNN was proposed by Moody and Darken \citeyearpar{moody_fast_1989} and Broomhhead and Lowe \citeyearpar{BroomhheadLowe1988} in the late of the 1980s for classification and function approximations. It was soon proved that GRBFNN is a universal approximator \citep{hornik_multilayer_1989, park_universal_1991, leshno_multilayer_1993} that can be arbitrarily close to a real-value function when the sufficient number of neurons is offered. The proof of universal approximability for GRBFNN can be interpreted as a process beginning with partitioning the domain of a target function into a grid, followed by using localized radial-basis functions to approximate the target function in each grid cell, and then aggregating the localized functions to globally approximate the target function. It is evident that this approach is not feasible for high-dimensional problems because it will lead to the exponential growth of neurons as the number of input dimensions increases. For example, $O(N^d)$ neurons will be required to approximate a $d$-variate function, with the domain of each dimension divided into $N$ segments.

To address this issue, researchers have heavily focused on selecting the optimal number of neurons as well as their centers and widths of GRBFNN such that the features of the target nonlinear map are well captured by the network. This has been mainly investigated through two strategies: (1) using supervised learning with dynamical adjustment of neurons (e.g., numbers, centers, and widths) according to the prescribed criteria and (2) performing unsupervised-learning-based preprocessing on input to estimate the placement and configuration of neurons.

For the former, Poggio and Girosi \citeyearpar{PoggioGirosi1990} as well as Wettschereck and Dietterich \citeyearpar{wettschereck1991} applied gradient descent to train generalized-radial-basis-function networks that have trainable centers. Regularization techniques \citep{PoggioGirosi1990} were adopted to maintain the parsimonious structure of GRBFNN. Platt \citeyearpar{PlattRAN1991} developed a two-layer network that dynamically allocates localized Gaussian neurons to the positions where the output pattern is not well represented. Chen et al. \citeyearpar{chen_orthogonal_1991} adopted an Orthogonal Least Square (OLS) method and introduced a procedure that iteratively selects the optimal centers that minimize the error reduction ratio until the desired accuracy is achieved. Huang et al. \citeyearpar{g-b_huang_generalized_2005} proposed a growing and pruning strategy to dynamically add/remove neurons based on their contributions to learning accuracy. 

The latter, unsupervised-learning-based preprocessing methods, have been more popular because it decouples the estimation of centers and widths from the computation of weights, which reduces the complexity of program as well as computational load. Moody and Darken \citeyearpar{moody_fast_1989} used the k-means clustering method \citep{WuK-mean2012} to determine the centers that minimize the Euclidean distance between the training set and centers, followed by the calculation of a uniform width by averaging the distance to the nearest-neighbor of all units. Carvalho and Brizzotti \citeyearpar{CarvalhoRBFClustering2001} investigated different clustering methods such as the iterative optimization (IO) technique, depth-first search (DF), and the combination of IO and DF for target recognition by RBFNNs. Niros and Tsekouras \citeyearpar{NirosfuzzyClustering2009} proposed a hierarchical fuzzy clustering method to estimate the number of neurons and trainable variables.

The optimization of widths has been of great interest more recently. Yao et al. \citeyearpar{Yaoetal2010} numerically observed that the optimal widths of radial basis function are affected by the spatial distribution of training data and the nonlinearity of approximated functions. With this in mind, they developed a method that determines the widths using the Euclidean distance between centers and second-order derivatives of a function. However, calculating the width of each neuron is computationally expensive. Instead of assigning each neuron a distinct width, it makes more sense to assign different widths to the neurons that represent different clusters for computational efficiency. Therefore, Yao et al. \citeyearpar{yaoetal2012} further proposed a method to optimize widths by dividing a global optimization problem into several subspace optimization problems that can be solved concurrently and then coordinated to converge to a global optimum. Similarly, Zhang et al. \citeyearpar{ZhangRFBFuzzy2019} introduced a two-stage fuzzy clustering method to split the input space into multiple overlapped regions that are then used to construct a local Gaussian-radial-basis-function network. 

However, the aforementioned methods all suffer from the curse of dimensionality. As the input dimension grows, the selection of optimal neurons itself can become cumbersome. To compound the problem, the number of optimal neurons can also rise exponentially when approximating high-dimensional and geometrically complex functions. Furthermore, the methods are designed for CPU-based, general-purpose computing machines but are not appropriate for tapping into the modern GPU-oriented machine-learning tools \citep{AbadiTensorflow2016,Paszke2019PyTorchAI} whose computational efficiency drop significantly when handling branching statements and dynamical memory allocation. This gap motivates us to reevaluate the structure of GRBFNN. As stated previously, the localized property of Gaussian functions is beneficial for identifying the parsimonious structure of GRBFNN with low input dimensions, but it also leads to the blow-up of the number of neurons in high dimensional situations. 

Given that the recent development of deep neural networks has shown promise in solving such problems, \textit{the main goal of this paper is to develop a deep-neural-network representation of GRBFNN such that it can be used for solving very high dimensional problems.} We approach this problem by utilizing the separable property of Gaussian radial basis functions. That is, every Gaussian-radial-basis function can be decomposed into the product of multiple uni-variate Gaussian functions. Based on this property, we construct a new neural network, namely separable Gaussian neural network (SGNN), whose number of layers is equal to the number of input dimensions, with the neurons of each layer formed by the corresponding uni-variate Gaussian functions. Through dividing the input into multiple columns by their dimensions and feeding them into the corresponding layers, the output equivalent to that of a GRBFNN is constructed from  multiplications and summations in the forward propagation. It should be noted that Poggio and Girosi \citeyearpar{PoggioGirosi1990} have reported the separable property of Gaussian-radial-basis functions and proposed using it for neurobiology even in 1990. 

SGNN offers several advantages.
\begin{itemize}
\item The number of neurons of SGNN is given by $O(dN)$ and increases linearly with the dimension of the input while the number of neurons of GRBFNN given by $O(N^d)$ grows exponentially. This reduction of neurons also decreases the number of trainable variables from $O(N^d)$ to $O(dN^2)$, yielding a more compact network than GRBFNN.
\item The reduction of trainable variables further decreases the computational load during training and testing of neural networks. As shown in Section \ref{SecSGNNvsGRBFNN}, this has led to 100 times speedup of training time for approximating tri-variate functions. 
\item SGNN is much easier to tune than other MLPs. Since the number of layers in SGNN is equal to the number of dimension of the input data, the only tunable network-structural hyper-parameter is the layer width, i.e. the number of neurons in a layer. This can significantly alleviate the tuning workload as compared to other MLPs that must simultaneously tune the width and depth of layers. 
\item SGNN holds a similar level of accuracy as GRBFNN, making it particularly suitable for approximating multi-variate functions with complex geometry. In Section \ref{SecCompareSGNN_DNNs}, it is shown that SGNN can yield three-order-of magnitude more accurate approximations for complex functions than MLPs with ReLU and Sigmoid functions.
\end{itemize} 

The rest of this paper is organized as follows. In Section \ref{SecSGNN}, we introduce the structure of SGNN and use it to approximate a multi-variate real-value function. In Section \ref{SecSGNNvsGRBFNN}, we compare SGNN and GRBFNN regarding the number of trainable variables and computational complexity of forward and backward propagation. In Section \ref{SecSubspaceGD}, we show that SGNN can preserve the dominant sub-eigenspace of the Hessian of GRBFNN in the gradient descent search. This property can help SGNN maintain a similar level of accuracy as GRBFNN while substantially improving computational efficiency. In Section \ref{SecNumResults}, we show the computational time of SGNN scales linearly with the increase of dimension and demonstrate its efficacy in function approximations through numerous examples. In Sections \ref{SecCompareSGNN_GRBFNN} and \ref{SecCompareSGNN_DNNs},  extensive comparisons between SGNN and GRBFNN and between SGNN and MLPs are performed. At last, the conclusion is summarized in Section \ref{SecConclusion}.

\section{Separable-Gaussian Neural Networks} 
\label{SecSGNN}

\begin{definition}
A $d$-variate function $f(x_1,x_2,\dots,x_d)$ is separable if it can be expressed as a product of multiple uni-variate functions; i.e.,
\begin{equation}
f(x_1,x_2,\dots,x_d)=f_1(x_1) \cdot f_2(x_2)\cdots  f_d(x_d).
\end{equation} 
\end{definition}

\begin{figure}[!ht]
\centering
\includegraphics[width=1\linewidth]{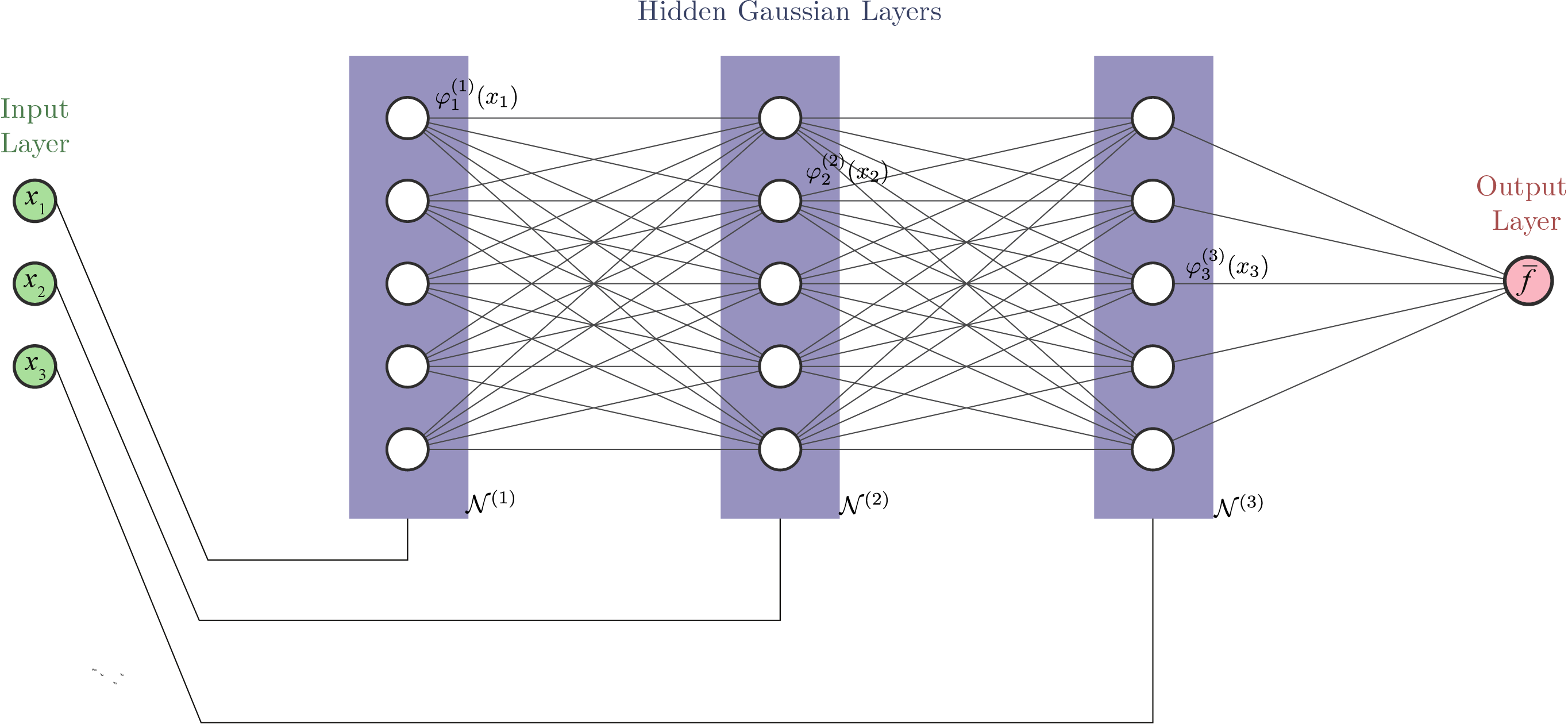}
\caption{The SGNN that approximates a tri-variate function. The input is divided and fed sequentially to each layer. Therefore, the depth (layers) of the NN is identical to the number of input dimensions. In this paper, the weights of the output layer are unity.}
\label{fig:gaussian-nn}
\end{figure}

\begin{remark}
The Gaussian radial-basis function
\begin{equation}
G(\mathbf{x})=\exp\left(-\sum_{i=1}^{d}\frac{(x_i-\mu_i)^2}{2\sigma_i^2}\right),
\label{eq:sec2:gaussian_radial_basis}
\end{equation} 
is separable and can be represented in the form
\begin{equation}
G(\mathbf{x})=\prod_{k=1}^{d}\varphi^{(k)}(x_k),
\label{eq:product_chain}
\end{equation}
where $\varphi^{(k)}(x_k)=\exp(-\frac{1}{2}(x_k-\mu_k)^2/\sigma_k^2)$, with $k=1,2,\dots,d$.
\end{remark}

The product chain in Eq. (\ref{eq:product_chain}) can be constructed through the forward propagation of a feedforward network with a single neuron per layer where $\varphi^{(k)}(x_k)$ is the neuron of the $k$-th layer. This way, the multi-variate Gaussian function $G(\mathbf{x})$ is reconstructed at the output of the network. By adding more neurons to each layer and assigning weights to all edges, we can eventually construct a network whose output is equivalent to the output of a GRBFNN. Fig. \ref{fig:gaussian-nn} shows an example of an SGNN approximating a tri-variate function. Next we use this property to define SGNN.

\begin{definition}
The separable-Gaussian neural network (SGNN) with $d$-dimensional input can be constructed in the form
\begin{align}
\mathcal{N}^{(0)}_i&=x_i,~1\leq i \leq d, \label{Eq: sec_4: layer 0}\\
\mathcal{N}^{(1)}_i&={\varphi}^{(1)}_i(x_1,{\mu}^{(1)}_i,{\sigma}^{(1)}_i), ~ 1\leq i \leq N_1, \label{Eq: sec_4: layer 1}\\
\mathcal{N}^{(\ell)}_i&={\varphi}^{(\ell)}_i(x_\ell,{\mu}^{(\ell)}_i,{\sigma}^{(\ell)}_i)\sum_{j=1}^{N_l} {W}^{(\ell)}_{ij}\mathcal{N}^{(\ell-1)}_j, ~~2\leq \ell\leq d, 1\leq i \leq N_l, \label{Eq:sec_4: layer l}\\
\bar{f}(\mathbf{x})&=\mathcal{N}(\mathbf{x})=\sum_{j=1} ^{N_d}\mathcal{N}^{(d)}_j, \label{Eq:sec_4: last layer}
\end{align}
where $N_i$ ($i=1,2,\dots, d$) represents the number of neurons of the $l$-th layer, $\mathcal{N}^{(l)}_i$ represents the output of the $i$-th Gaussian neuron (activation function) of the $l$-th layer.
\end{definition}

Substitution of Eqs. (\ref{Eq: sec_4: layer 0}) to (\ref{Eq:sec_4: layer l}) into Eq. (\ref{Eq:sec_4: last layer}) yields
\begin{equation}
\bar{f}(\mathbf{x})=\sum_{i_{d}=1}^{N_d}\sum_{i_{d-1}=1}^{N_{d-1}}\dots\sum_{i_1=1}^{N_1}\left[W^{(d-1)}_{i_di_{d-1}}W^{(d-2)}_{i_{d-1}i_{d-2}}\dots W^{(1)}_{i_2i_{1}}\right]\prod_{\ell=1}^{d}{\varphi}^{(\ell)}_{i_\ell}(x_\ell),
\label{Eq:sum_multi_layer}
\end{equation}
with 
\begin{equation}
\prod_{\ell=1}^{d}{\varphi}^{(\ell)}_{i_\ell}(x_\ell) = {\varphi}^{(d)}_{i_d}(x_d){\varphi}^{(d-1)}_{i_{d-1}}(x_{d-1})\dots{
\varphi}^{(1)}_{i_1}(x_1),
\end{equation}
where $W^{(l)}_{i_{l+1}i_{l}}$ ($l=1,2,\dots,d$) represents the weight of the $i_{(l+1)}$-th neuron of the $(l+1)$-th layer and the $i_{l}$-th neuron of the $l$-th layer. The loss function of the SGNN is defined in the form
\begin{equation}
J= \lVert f-\bar{f} \rVert_2=\sqrt{\sum_{i=1}^{d}\left[f(x_i)-\bar{f}(x_i)\right]^2}.
\end{equation}

The center $\mu_i^{(l)}$ and width $\sigma_i^{(l)}$ in the Gaussian function $\varphi_i^{(l)}$ can also be treated as trainable. They are not included in this discussion for simplicity.

\section{SGNN vs. GRBFNN}
\label{SecSGNNvsGRBFNN}
Without loss of generality, the analysis below will assume that each hidden layer has $N$ neurons. To understand how the weights of SGNN relate to those of GRBFNN, we equate Eqs. (\ref{eq:GRBFNN}) and (\ref{Eq:sum_multi_layer}), which yields a nonlinear map 
\begin{equation}
\mathcal{W}_j=g_j\left(W^{(d-1)}_{i_di_{d-1}},W^{(d-2)}_{i_{d-1}i_{d-2}},\dots,W^{(1)}_{i_2i_{1}}\right),
\label{eq:mapping_w_to_W}
\end{equation}
whose explicit form is
\begin{equation}
\mathcal{W}_j = W^{(d-1)}_{i_di_{d-1}}W^{(d-2)}_{i_{d-1}i_{d-2}}\dots W^{(1)}_{i_2i_{1}},
\label{Eq:wegiht_mapping}
\end{equation}
with
\begin{equation}
j=i_1+i_2N+\dots+i_d N^{d-1}.
\end{equation}
It is evident that SGNN can be transformed into GRBFNN. However, GRBFNN can be converted into SGNN if and only if the mapping of Eq. (\ref{Eq:wegiht_mapping}) is invertible. 

Because the mapping of Eq. (\ref{Eq:wegiht_mapping}) is not uniquely invertible, it is difficult to prove the universal approximability of SGNN. However, this paper will present extensive numerical experiments to show that SGNN can achieve comparable (occasionally even greater) accuracy with much less computation effort than GRBFNN.
In addition, SGNN can have superior performance in approximating complex functions than deep neural networks with activation functions such as ReLU and Sigmoid as shown in Section \ref{SecCompareSGNN_DNNs}. 

In the following, we demonstrate the computational efficiency of SGNN over GRBFNN in terms of trainable variables and the number of floating-point operations of forward and backward propagation. 

\subsection{Trainable Variables}
Let us now treat the center and width of the uni-variate Gaussian function in SGNN as trainable. The total number $N_t$ of trainable variables of SGNN is given by
\begin{equation}
N_t=\begin{cases}
N + 2N & \mathbf{x} \in \mathbb{R}^1,\\
(d-1)N^2+2dN & \mathbf{x} \in \mathbb{R}^d, \text{ for } d \geq 2.
\end{cases}
\end{equation}
Note that the number of trainable variables of GRBFNN is $N^d$, identical to its number of neurons. SGNN and GRBFNN have identical weights when the number of layers is smaller than or equal to two. In other words, they are mutually convertible and the mapping of Eq. (\ref{Eq:wegiht_mapping}) is invertible when $d\leq 2$. However, for high-dimensional problems, as shown in Table \ref{table:varibles_of_SGNN_vs_GRBFNN}, SGNN can substantially reduce the number of trainable variables, making it more tractable than GRBFNN.

\begin{table}[!ht]
\centering
\caption{Neurons and trainable variables of SGNN and GRBFNN.}
 \vspace{5mm}
\begin{tabular}{c|c|c}
\toprule
& Neurons & No. of variables\\
\hline
SGNN & $O(dN)$ & $O(dN^2)$ \\
GRBFNN & $O(N^d)$ & $O(N^d)$ \\
\bottomrule
\end{tabular}
\label{table:varibles_of_SGNN_vs_GRBFNN}
\end{table}

\subsection{Forward Propagation}
Assume the size of the input dataset is $m$. Using Eqs. (\ref{Eq: sec_4: layer 0}) to (\ref{Eq:sec_4: last layer}), we can estimate the number of floating-point operations (FLOP) of the forward pass in SGNN. More specifically, the number of FLOP to calculate the output of the $k$-th layer with the input from the previous layer is
\begin{equation}
FLOP^{(k)}(\mathcal{N}(\mathbf{x})) = m(2N^2 + 6N), \text{ for } 2 \leq k\leq d,
\end{equation}
where $2N^2$ is the number of arithmetic operations by the product of weights and Gaussian functions of the $k$-th layer, and $6N$ is the number of calculations for Gaussian functions of the layer, and $m$ is the size of input dataset. In addition, the number of FLOP associated with the first and output layer are 
\begin{eqnarray}
FLOP^{(1)}(\mathcal{N}(\mathbf{x}))= 6mN,\\
FLOP^{(d+1)}(\mathcal{N}(\mathbf{x}))= mN.
\end{eqnarray}
Therefore, the total number of FLOP is 
\begin{equation}
O(FLOP_{fp})=O\left(\sum_{i=1}^{d+1}FLOP^{(i)}\right)= O(mdN^2).
\end{equation}

The number of operations increases linearly with the increase of the number of layers or the dimension of the input vector $d$. On the other hand, the computational complexity of FLOP of RBGNN is 
\begin{equation}
O(FLOP_{\widetilde{fp}})=O(mN^d),
\end{equation}
regardless of the trainability of the centers and width of Gaussian functions.

\subsection{Backward Propagation} 
Accurately estimating the computational complexity of backward propagation is challenging because techniques such as auto differentiation \citep{BaydinetalAutoDiff2018} and computational graphs \citep{AbadiTensorflow2016} have optimized the original mathematical operations for improving the performance. Auto differentiation evaluates the derivative of numerical functions using dual numbers with the chain rule broken into a sequence of operations such as addition, multiplication, and composition. During forward propagation, intermediate values in computational graphs are recorded for backward propagation.

We analyze the operations of backward propagation with respect to a single neuron of the $l$-th layer. The partial derivatives of $\bar{f}(\mathbf{x})$ with respect to $\mathcal{W}^{(l)}_{j}$, ${ \mu^{(l)}_{j}}$, and ${ \sigma^{(l)}_{j}}$ of the $l$-th ($1 \leq l \leq d$) layer in SGNN are 



\begin{align}
\frac{\partial\bar{f}}{\partial \mathcal{W}^{(l)}_{j}}&=\left[\frac{\partial\bar{f}}{\partial\mathcal{N}^{(l+1)}_j}\right]^T\frac{\partial \mathcal{N}^{(l+1)}_j}{\partial \mathcal{W}^{(l)}_{j}},\\
\frac{\partial\bar{f}}{\partial \mu^{(l)}_{j}}&=\left[\frac{\partial\bar{f}}{\partial\mathcal{N}^{(l+1)}_j}\right]^T\frac{\partial \mathcal{N}^{(l+1)}_j}{\partial \mu^{(l)}_{j}},\\
\frac{\partial\bar{f}}{\partial \sigma^{(l)}_{j}}&=\left[\frac{\partial\bar{f}}{\partial\mathcal{N}^{(l+1)}_j}\right]^T\frac{\partial \mathcal{N}^{(l+1)}_j}{\partial \sigma^{(l)}_{j}},
\end{align}
with
\begin{equation}
\left[\frac{\partial\bar{f}}{\partial\mathcal{N}^{(l+1)}_j}\right]=\left[\frac{\partial\bar{f}}{\partial \mathcal{N}^{(l+2)}}\right]^T\left[\frac{\partial \mathcal{N}^{(l+2)}}{\partial\mathcal{N}^{(l+1)}_j}\right],
\label{eq:sec4.3_partial_derivative_output_layer_l+1}
\end{equation}
where
\begin{equation}
\mathcal{N}^{(l+2)} = ({\mathcal{N}}_1^{(l+2)}, {\mathcal{N}}_2^{(l+2)},\dots, {\mathcal{N}}_N^{(l+2)})^T.
\end{equation}
 The backward prorogation with respect to the $j$-th neuron of the $l$-th ($1\leq l \leq n-1$) layer can be divided into three steps:
 
\begin{enumerate}
\item Compute the gradient of $\bar{f}$ with respect to the output of the $1$-st neuron in the $(l+1)$-th layer, $\mathcal{N}_j^{(l+1)}$, as shown in Eq. (\ref{eq:sec4.3_partial_derivative_output_layer_l+1}), where $\left[\frac{\partial\bar{f}}{\partial \mathcal{N}^{(l+2)}
}\right]^T$ can be accessed from the back propagation of the $(l+2)$-th layer. This leads to $2N$ FLOP due to the dot product of two vectors. 
\item Calculate the partial derivatives of $\mathcal{N}_j^{(l+1)}$ with respect to weights, center, and width. Since the calculation of derivatives is computationally cheap, the analysis below will neglect the operations used to evaluate derivatives. This shall not affect the conclusion.
\item Propagate the gradients backward. This produces $N+2$ operations.
\end{enumerate} 

Therefore, the number of FLOP of the $l$-th layer is approximately $m(3N^2+2N)$, where $m$ is the volume of the input dataset. The backward propagation of the last layer leads to $N$ operations. In total, the number of FLOP by backward propagation is 
\begin{equation}
O(FLOP_{bp})=O(mdN^2).
\end{equation}
On the other hand, the backward propagation FLOP number of GRBFNN is 
\begin{equation}
O(FLOP_{\widetilde{bp}})=O(mN^d).
\end{equation}

\section{Subspace Gradient Descent}
\label{SecSubspaceGD}

As illustrated in Section \ref{SecSGNNvsGRBFNN}, SGNN has exponentially fewer trainable variables than the associated GRBFNN for high-dimensional input. In other words, GRBFNN may be over-parameterized. The recent work \citep{sagun2017eigenvalues,sagun2018empirical,gurari2018gradient} has shown that optimizing a loss function constructed by an over-parameterized neural network can lead to Hessian matrices that possess few dominant eigenvalues with many near-zero ones before and after training. This means the gradient descent can happen in a small subspace. Inspired by their work, we consider infinitesimal variation of the loss function $J$ for GRBFNN a
\begin{equation}
dJ=\left[\frac{\partial J}{\partial \tilde{\boldsymbol{\theta}}}\right]^Td\tilde{\boldsymbol{\theta}}+\frac{1}{2}d\tilde{\boldsymbol{\theta}}^T\tilde{H}d\tilde{\boldsymbol{\theta}}+h.o.t.(\lVert d\tilde{\boldsymbol{\theta}}\rVert^3),
\label{eq:sec_6_dJ}
\end{equation}
where $\tilde{\boldsymbol{\theta}}$ represents a vector of all trainable weights, and
\begin{equation}
\tilde{\mathbf{H}}=\frac{\partial^2 J}{\partial \tilde{\boldsymbol{\theta}}^T\tilde{\boldsymbol{\theta}}} ,
\end{equation}
is the associated Hessian matrix. The centers and widths of Gaussian functions are assumed to be constant for simplicity. Since the Hessian matrix $\tilde{H}$ is symmetric, we can represent it in the form
\begin{equation}
\tilde{\mathbf{H}}=\mathbf{P}^T\begin{pmatrix}
\boldsymbol{\lambda}_d & \mathbf{0}\\
\mathbf{0} & \boldsymbol{\lambda}_s
\end{pmatrix}\mathbf{P},
\end{equation}
where $\boldsymbol{\lambda}_d=\mathrm{diag}(\lambda_1, \lambda_2,\dots,\lambda_k, \dots, \lambda_{dN}) $ are the $k$ dominant eigenvalues padded by $(dN-k)$ non-dominant ones (assuming $k<dN$), and $\boldsymbol{\lambda}_s$ = $\mathrm{diag}(\lambda_{dN+1},\lambda_{dN+2},\dots,\lambda_{N^d})$ are the rest non-dominant eigenvalues.

Let ${\boldsymbol{\theta}}$ be the weights of SGNN. The variation of the mapping from ${\boldsymbol{\theta}}$ to $\tilde{\boldsymbol{\theta}}$ in Eq. (\ref{eq:mapping_w_to_W}) reads 
\begin{equation}
d{\tilde{\boldsymbol{\theta}}} = \frac{\partial \mathbf{g}}{\partial {\boldsymbol{\theta}}}d{\boldsymbol{\theta}},
\label{eq: sec_6_jacobian_mapping}
\end{equation}
where $\frac{\partial \mathbf{g}}{\partial {\boldsymbol{\theta}}}$: $\mathbb{R}^{dN}\mapsto \mathbb{R}^{N^d \times dN}$. It should be noted that $\frac{\partial \mathbf{g}}{\partial {\boldsymbol{\theta}}}$ is a super sparse matrix.

Substitution of Eq. (\ref{eq: sec_6_jacobian_mapping}) into Eq. (\ref{eq:sec_6_dJ}) 
\begin{equation}
dJ=\left[\frac{\partial J}{\partial \tilde{\boldsymbol{\theta}}}\right]^T\frac{\partial \mathbf{g}}{\partial {\boldsymbol{\theta}}}d{\boldsymbol{\theta}}
+\frac{1}{2}d{\boldsymbol{\theta}}^T \mathbf{H} d{\boldsymbol{\theta}}
+h.o.t.(\lVert d\tilde{\boldsymbol{\theta}}\rVert^3),
\end{equation}
with
\begin{equation}
\begin{split}
\mathbf{H}&= \left[\frac{\partial \mathbf{g}}{\partial {\boldsymbol{\theta}}}\right]^T
\tilde{\mathbf{H}}
\frac{\partial \mathbf{g}}{\partial {\boldsymbol{\theta}}}\\
&=\left[\frac{\partial \mathbf{g}}{\partial {\boldsymbol{\theta}}}\right]^T \mathbf{P}^T
\begin{pmatrix}
\boldsymbol{\lambda}_d & \mathbf{0}\\
\mathbf{0} & \boldsymbol{\lambda}_s
\end{pmatrix}  \mathbf{P}
\left[ \frac{\partial \mathbf{g}}{\partial {\boldsymbol{\theta}}} \right].
\end{split}
\label{eq:subspace_H_bar}
\end{equation}
Let
\begin{equation}
\begin{pmatrix}
\mathbf{Q}_{d}\\
\mathbf{Q}_{s}
\end{pmatrix}= \mathbf{P}
\left[ \frac{\partial \mathbf{g}}{\partial {\boldsymbol{\theta}}} \right],
\label{eq:subspace_Q_matrix}
\end{equation}
where $\mathbf{Q}_{d}\in \mathbb{R}^{dN\times dN}$ and $\mathbf{Q}_{s}\in \mathbb{R}^{(N^d-dN)\times dN}$.
Substitution Eq. (\ref{eq:subspace_Q_matrix}) into Eq. (\ref{eq:subspace_H_bar}) yields
\begin{equation}
\begin{split}
\mathbf{H} &= \mathbf{Q}_d^T\boldsymbol{\lambda}_d \mathbf{Q}_d+\mathbf{Q}_s^T\boldsymbol{\lambda}_s \mathbf{Q}_s\approx \mathbf{Q}_d^T\boldsymbol{\lambda}_d \mathbf{Q}_d.
\end{split}
\label{Eq:H_SGNN}
\end{equation}

Therefore, the dominant eigenvalues of the Hessian of GRBFNN are also included in the corresponding SGNN. This means that the gradient of SGNN can descend in the mapped dominant non-flat subspace of GRBFNN, which may contribute to the comparable accuracy and training efficiency of SGNN as opposed to GRBFNN, as discussed in Section \ref{SecSGNNvsGRBFNN}.

\section{Numerical Experiments}
\label{SecNumResults}

\subsection{Candidate Functions}
We consider ten candidate functions from \citep{andras_function_2014,Andras2018} and modified them, as listed in Table \ref{table:candidate_functions}. The functions cover a range of distinct features, including sinks, sources, flat and s-shaped surfaces, and multiple sinks and sources, which can assist in benchmarking  function approximations of different neural networks. 

We generate uniformly distributed sample sets to train neural networks for each run, with upper and lower bounds of each dimension ranging from -8 to 8. During the training process, we employ mini-batch gradient descent with the optimizer Adam in Tensorflow to update model parameters. The optimizer uses its default training parameters and stops if no improvement of loss values is achieved in four consecutive epochs. The dataset is divided into a training set comprising 80\% of the data and a validation set consisting of the remaining 20\%. The mini-batch size, number of neurons, and data points are selected to balance the convergence speed and accuracy. All tests are performed on a Windows-10 desktop with a 3.6HZ, 8-core, Intel i7-9700K CPU and a 64GB Samsung DDR-3 RAM.
\begin{table}[!ht]
\centering
\begin{threeparttable}
\def\arraystretch{1.4}
\caption{Candidate functions and their features.}
\label{table:candidate_functions}
\vspace{5mm}
\footnotesize{
\begin{tabular}{lll}
\toprule
Functions & Features& Explicit expression\\
\hline
Root sum squared & Sink & $
f_1(\mathbf{x}) = \left(\sum_{i=1}^{d}x_i^2\right)^{\frac{1}{2}}$\\
Second-degree polynomial & Saddle & $f_2(\mathbf{x})=\frac{1}{50}\sum_{j=1}^{d}x_j^2x_{j+1}$\\
Exponential-square sum & Flatter sink & $f_3(\mathbf{x})=\frac{1}{5}\sum_{j=1}^{d}e^{x_j^2/50}$\\
Exponential-sinusoid sum & Sink \& Source & $ f_4(\mathbf{x})=\frac{1}{5}\sum_{j=1}^{d}e^{x_j^2/50}\sin(y_{j})$\\
Polynomial-sinusoid sum & Sink \& Source & $ f_5(\mathbf{x})=\frac{1}{50}\sum_{j=1}^{d}x_j^2\cos(j*x_j)$\\
Inverse-exponential-square sum & Source & $f_6(\mathbf{x})=10/\sum_{j=1}^{d}e^{x_j^2/25}$\\
Sigmoidal & S-shaped surface & $f_7(\mathbf{x})=10/(1+e^{-\frac{1}{5}\sum_{i=1}^{d}x_j})$\\
Gaussian & Flatter source & $ f_8(\mathbf{x})=10e^{-\frac{1}{100}\sum_{j=1}^{5}x_j^2}$\\
Linear & Flat & $f_9(\mathbf{x})=\sum_{j=1}^{d}x_j$\\
Constant & Flat & $f_{10}(\mathbf{x})=1$\\
\bottomrule
\end{tabular}
}
 \begin{tablenotes}
{\footnotesize
\item Note: In $f_4$, $y_j=x_{j+1}$ with $j=1,2,\dots, d-1$ and $y_d=x_1$.}
\end{tablenotes}

\end{threeparttable}
\end{table}

\subsection{Dimension Scalability}
In order to understand the dimensional scalability of SGNN, we applied SGNN to candidate functions with the number of dimensions from two to five. For comparison, the data points were kept as 16384 such that sufficient data was sampled for 5-D functions, i.e. $d=5$. Each layer has fixed 20 uni-variate Gaussian neurons, with initial centers evenly distributed in each dimension and widths being the distance between two adjacent centers. The training time per epoch grows linearly as the dimension increases, with an increment of 0.02 seconds/epoch per layer. For the majority of candidate functions, SGNN can achieve the accuracy level of $10^{-4}$.  It is sufficient to approximate the 5-D functions by SGNN with 5 layers and in total 100 neurons. The configuration of SGNN cannot well approximate the function $f_5$ in 4-D. This can be easily resolved by adding more neurons to the neural network (see a similar example in Table \ref{table:SGG_vs_ReLU_configuration}). In summary, the computational time of SGNN scales linearly with the increase of dimensions.

\begin{table}[!ht]
\caption{The computation time per epoch of SGNN scales linearly with the increase of dimensions. Data is generated by averaging the results of 30 runs. Data Size: 16384, Mini-batch size: 256. Neurons per layer: 20.}
\label{table:CiteMe}
\vspace{5mm}
\centering
\footnotesize{
\begin{tabular}{c|cc|cc|cc|cc}
\toprule
& \multicolumn{2}{c|}{2D} & \multicolumn{2}{c|}{3D} & \multicolumn{2}{c|}{4D} & \multicolumn{2}{c}{5D}\\
\hline
& sec/epoch & Loss & sec/epoch & loss& sec/epoch & loss & sec/epoch & loss\\
\hline 
$f_1$ & \textbf{0.065} & 2.31E-04 & \textbf{0.081} & 4.43E-04 & \textbf{0.099} & 1.41E-03 & \textbf{0.113} & 4.43E-04 \\
$f_2$ & \textbf{0.063} & 7.34E-05 & \textbf{0.081} & 8.31E-04 & \textbf{0.098} & 2.50E-03 & \textbf{0.114} & 8.31E-04 \\
$f_3$ & \textbf{0.064} & 4.26E-06 & \textbf{0.083} & 1.50E-05 & \textbf{0.106} & 4.13E-05 & \textbf{0.110} & 1.50E-05 \\
$f_4$ & \textbf{0.065} & 2.80E-06 & \textbf{0.084} & 2.91E-05 & \textbf{0.100} & 9.12E-05 & \textbf{0.108} & 2.91E-05 \\
$f_5$ & \textbf{0.063} & 7.40E-05 & \textbf{0.083} & 7.53E-04 & \textbf{0.099} & 1.00E-01 & \textbf{0.107} & 7.53E-04 \\
$f_6$ & \textbf{0.063} & 1.39E-06 & \textbf{0.083} & 1.27E-05 & \textbf{0.101} & 2.11E-05 & \textbf{0.115} & 1.27E-05 \\
$f_7$ & \textbf{0.063} & 4.44E-05 & \textbf{0.083} & 4.08E-04 & \textbf{0.099} & 1.89E-03 & \textbf{0.111} & 4.08E-04 \\
$f_8$ & \textbf{0.063} & 1.97E-05 & \textbf{0.083} & 4.36E-05 & \textbf{0.100} & 7.76E-05 & \textbf{0.113} & 4.36E-05 \\
$f_9$ & \textbf{0.063} & 2.27E-04 & \textbf{0.079} & 1.76E-03 & \textbf{0.099} & 9.93E-03 & \textbf{0.111} & 1.76E-03 \\
$f_{10}$ & \textbf{0.064} & 3.51E-06 & \textbf{0.082} & 6.32E-06 & \textbf{0.101} & 9.84E-06 & \textbf{0.113} & 6.32E-06 \\
\bottomrule
\end{tabular}
}
\end{table}

\begin{figure}[!ht]
\centering
\includegraphics[width=\linewidth]{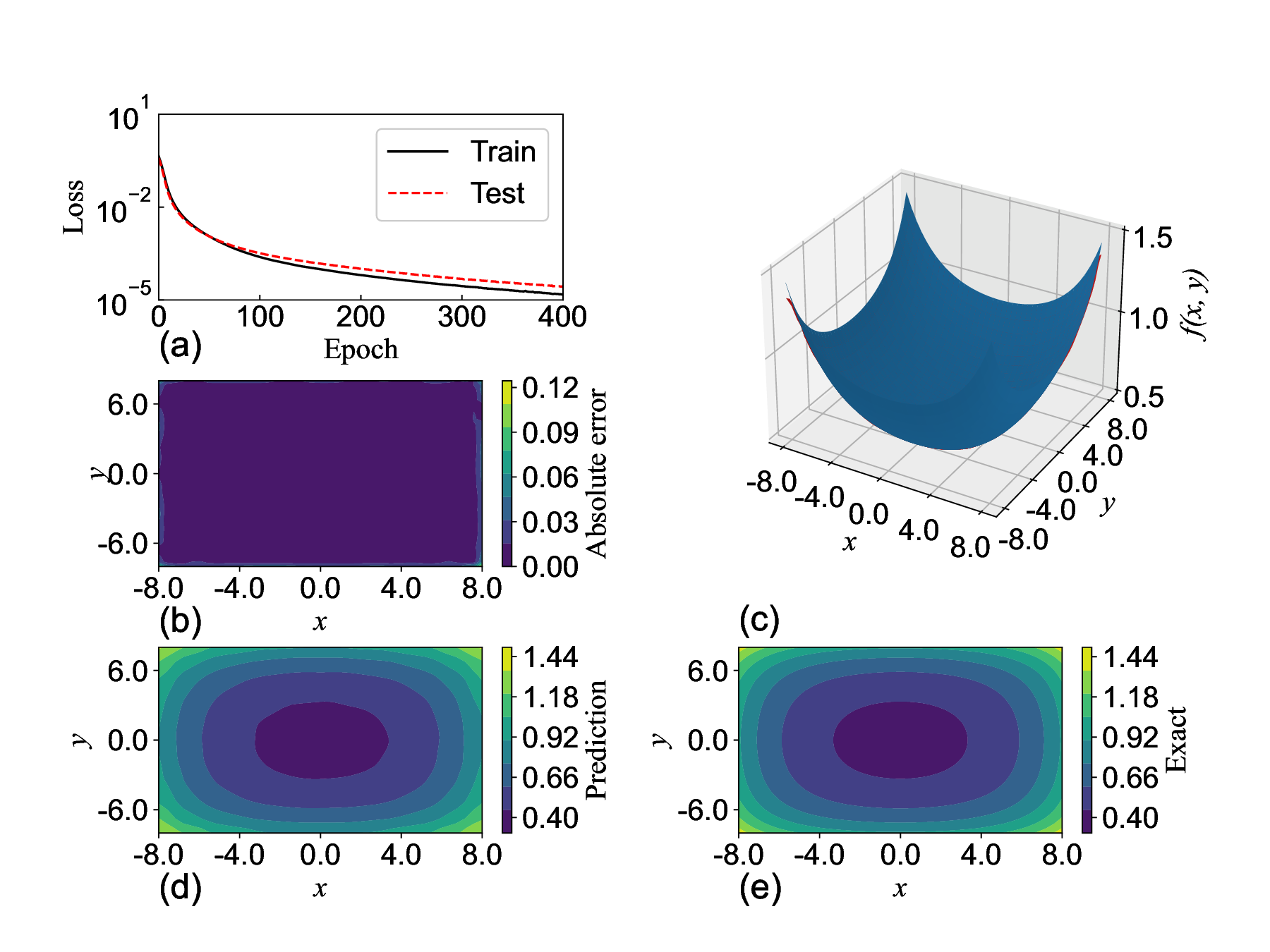}
\caption{Approximating the two-dimensional function $f_3$ by SGNN. (a) Training history, (b) absolute error, (c) prediction vs. exact value, (d) prediction, (e) ground truth. Size of training dataset: 2048.}
\label{fig:gaussianexpsquresum2023-02-21}
\end{figure}

Next, two- and five-dimensional examples are selected to illustrate the expressiveness of SGNN in function approximations. The number of neurons, training size, and mini-batch size are fine-tuned to achieve optimal results. 

\begin{figure}[!ht]
\centering
\includegraphics[width=\linewidth]{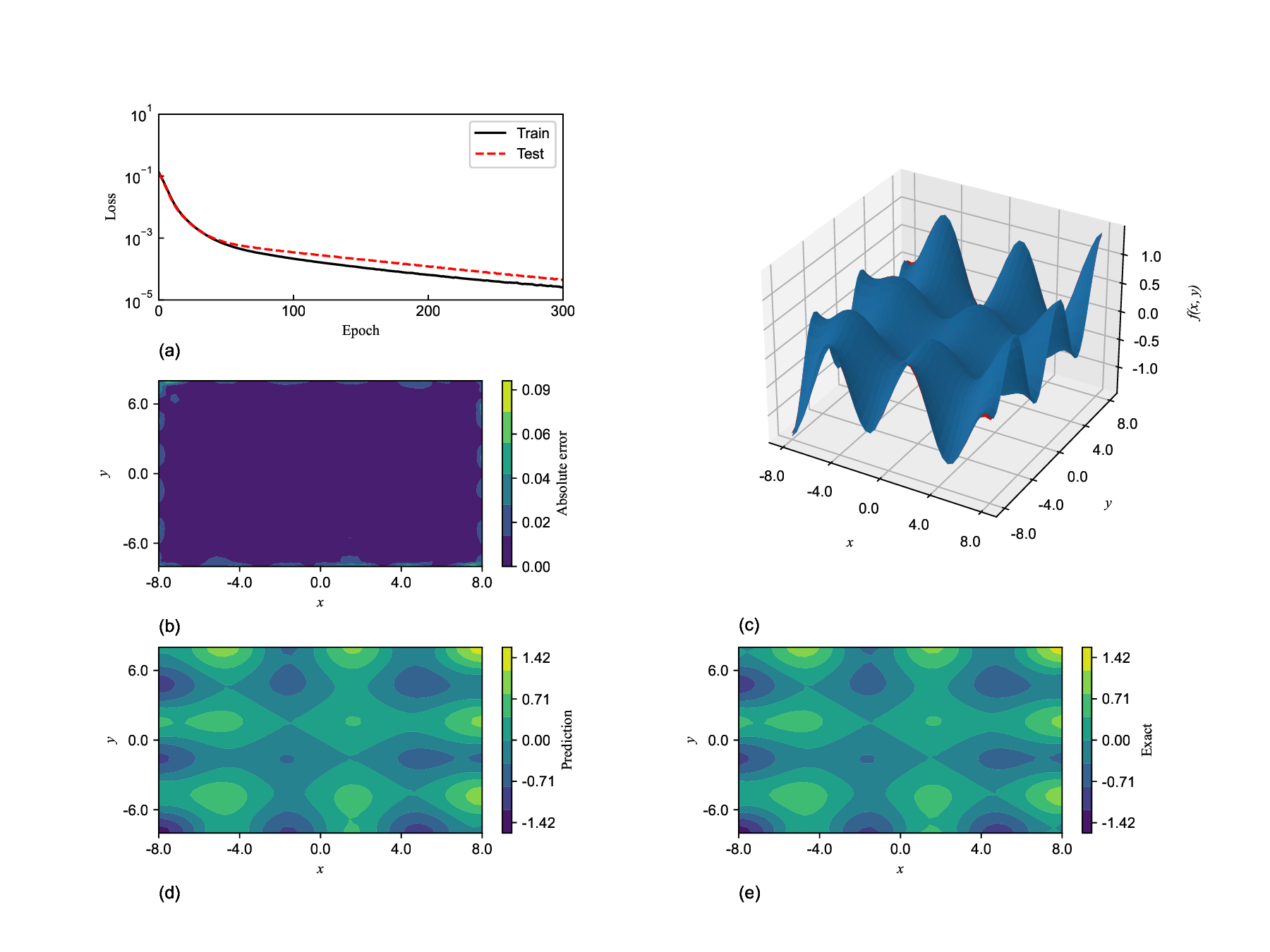}
\caption{Approximating the two-dimensional $f_4$ by SGNN. (a) Training history, (b) absolute error, (c) prediction vs. exactness, (d) prediction, (e) ground truth. Size of training dataset: 2048.}
\label{fig:gaussianexposinsum2023-02-21}
\end{figure}

\subsection{2-D Examples}
First, SGNN is used to approximate the two-dimension function $f_3(\mathbf{x})=1/5e^{(x_1^2+x_2^2)/50}$, which has four sharp peaks and one flat valley in the domain. As illustrated in Fig. \ref{fig:gaussianexpsquresum2023-02-21}(a), the optimizer converges in 400 steps, with the difference between training and test sets at the magnitude level of $10^{-4}$. Figs. \ref{fig:gaussianexpsquresum2023-02-21}(b)-(e) show that the prediction by SGNN is nearly identical to the ground truth, except for the domain near boundaries. This can be attributed to fewer sampling points in the neighborhood of the boundaries. Better alignment can be achieved by sampling extra boundary points to the input dataset.

SGNN maintains its level of accuracy as candidate functions become more complex. For example, Fig. \ref{fig:gaussianexposinsum2023-02-21} presents the approximation of $f_4(\mathbf{x})=\frac{1}{5}(e^{x_1^2/50}\sin x_2+e^{x_2^2/50}\sin x_1)$. SGNN can approximate $f_4$ with the same level of accuracy as $f_3$ even with fewer training epochs, possibly led by the localization property of Gaussian function. The largest error again appears near boundaries, with a percentage error less than 8\%. Inside the domain, the computed values precisely match the exact ones. As visualized in Figs. \ref{fig:gaussianexposinsum2023-02-21}(d) and (e), the prediction by SGNN can fully capture features of the function. 

\begin{figure}[!ht]
\centering
\includegraphics[width=\linewidth]{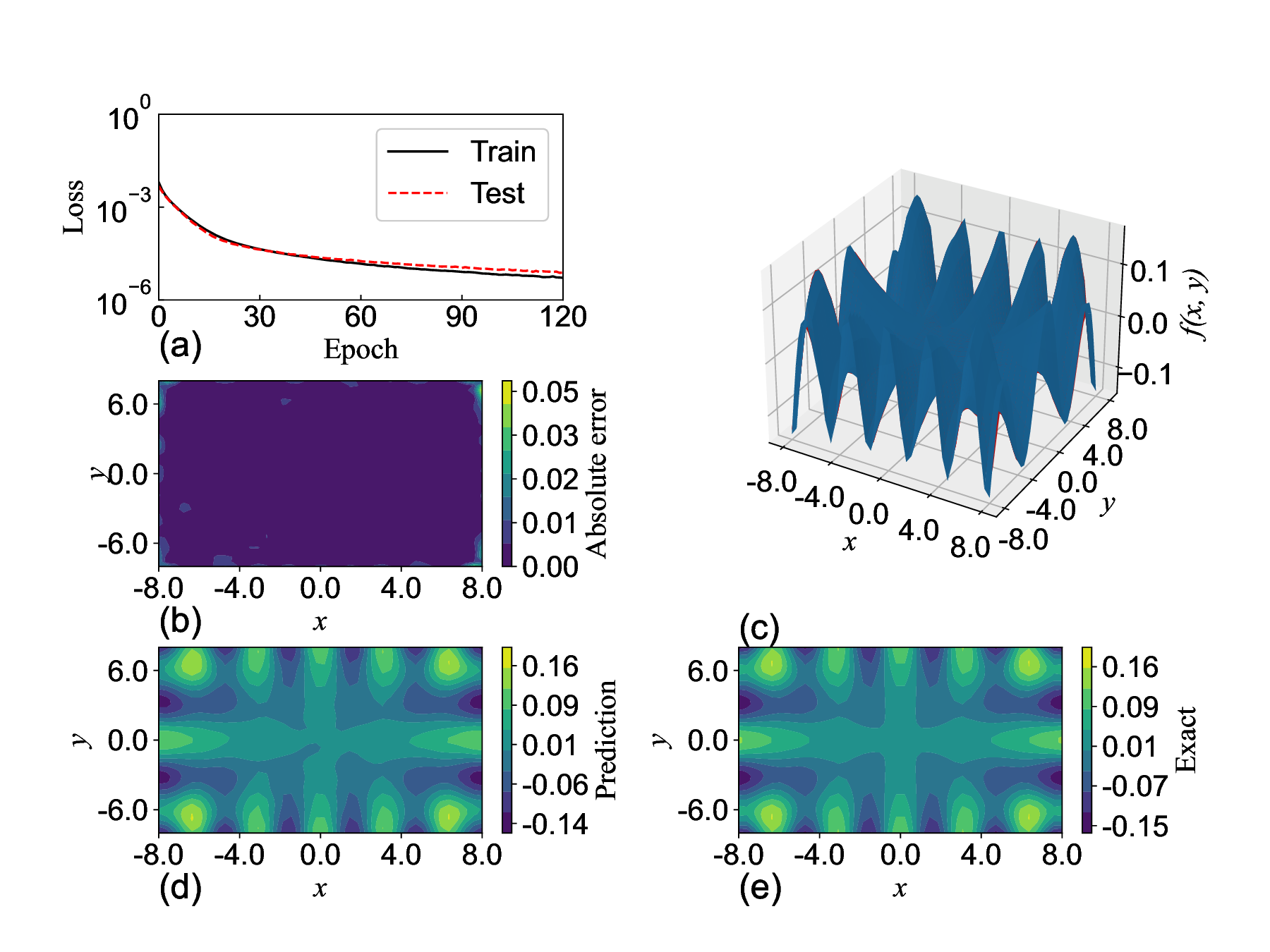}
\caption{Approximating the two-dimensional $f_5$ by SGNN. (a) Training history, (b) absolute error, (c) prediction vs. exactness, (d) 2-D projection of prediction, (e) 2-D projection of ground truth. Size of training dataset: 2048.}
\label{fig:gaussian2023-02-10-polysinsum}
\end{figure}

This finding is corroborated in Fig. \ref{fig:gaussian2023-02-10-polysinsum}, which presents the approximation of the function $f_5(\mathbf{x})=\frac{1}{50}(x_1^2\cos x_1+x_2^2\cos 2x_2)$ by SGNN. The function, different from $f_4$, possesses peaks and valleys near boundaries and becomes flat in the vicinity of origin, as illustrated in Figs. \ref{fig:gaussian2023-02-10-polysinsum}(c)-(e). Interestingly, the neural network converges faster than the network for $f_4$. This indicates that the loss function may become more convex and contain fewer flat regions. One possible reason is that as the function becomes more complex, more Gaussian neurons are active and have larger weights, increasing the loss gradients. The largest error is again observed near boundaries. As shown in Fig. \ref{fig:gaussian2023-02-10-polysinsum}, SGNN can well capture the features of the target function $f_5$. Due to the gradient configuration of the color bar, the small offset with respect to the ground truth occurs near the origin, but the corresponding absolute errors are very small, as shown in Fig. \ref{fig:gaussian2023-02-10-polysinsum}(b).

\subsection{5-D Examples}
\begin{figure}[!ht]
\centering
\includegraphics[width=1\linewidth]{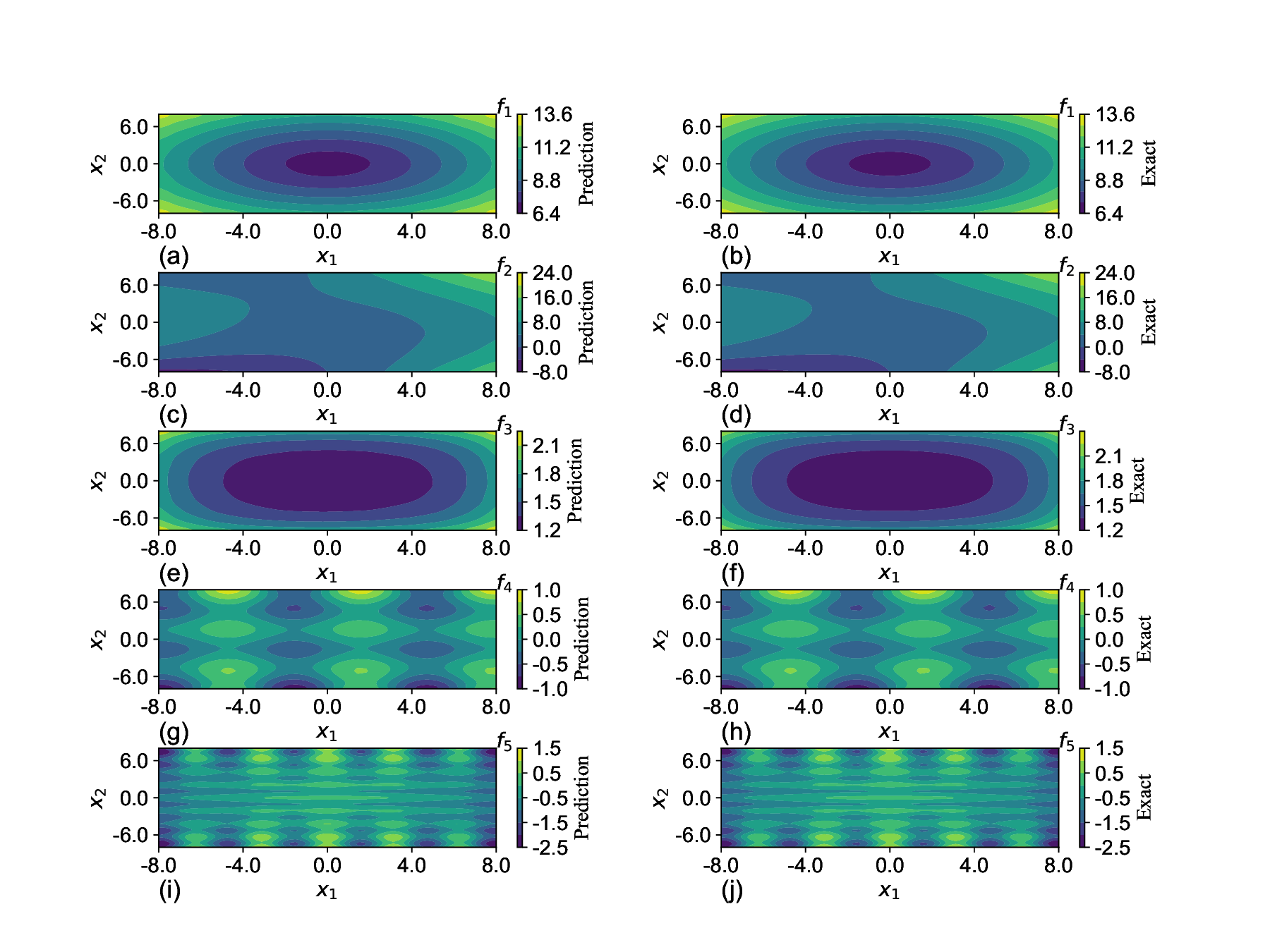}
\caption{Prediction vs. exact of $f_1$-$f_5$ in five dimensions. The plots are generated by projecting the surface to $x_1-x_2$ plane with other coordinates fixed to zero. The left panel is for prediction; the right panel is for the exact value. Size of training dataset: 32768.}
\label{fig:f1tof5}
\end{figure}

\begin{figure}[!ht]
\centering
\includegraphics[width=1.0\linewidth]{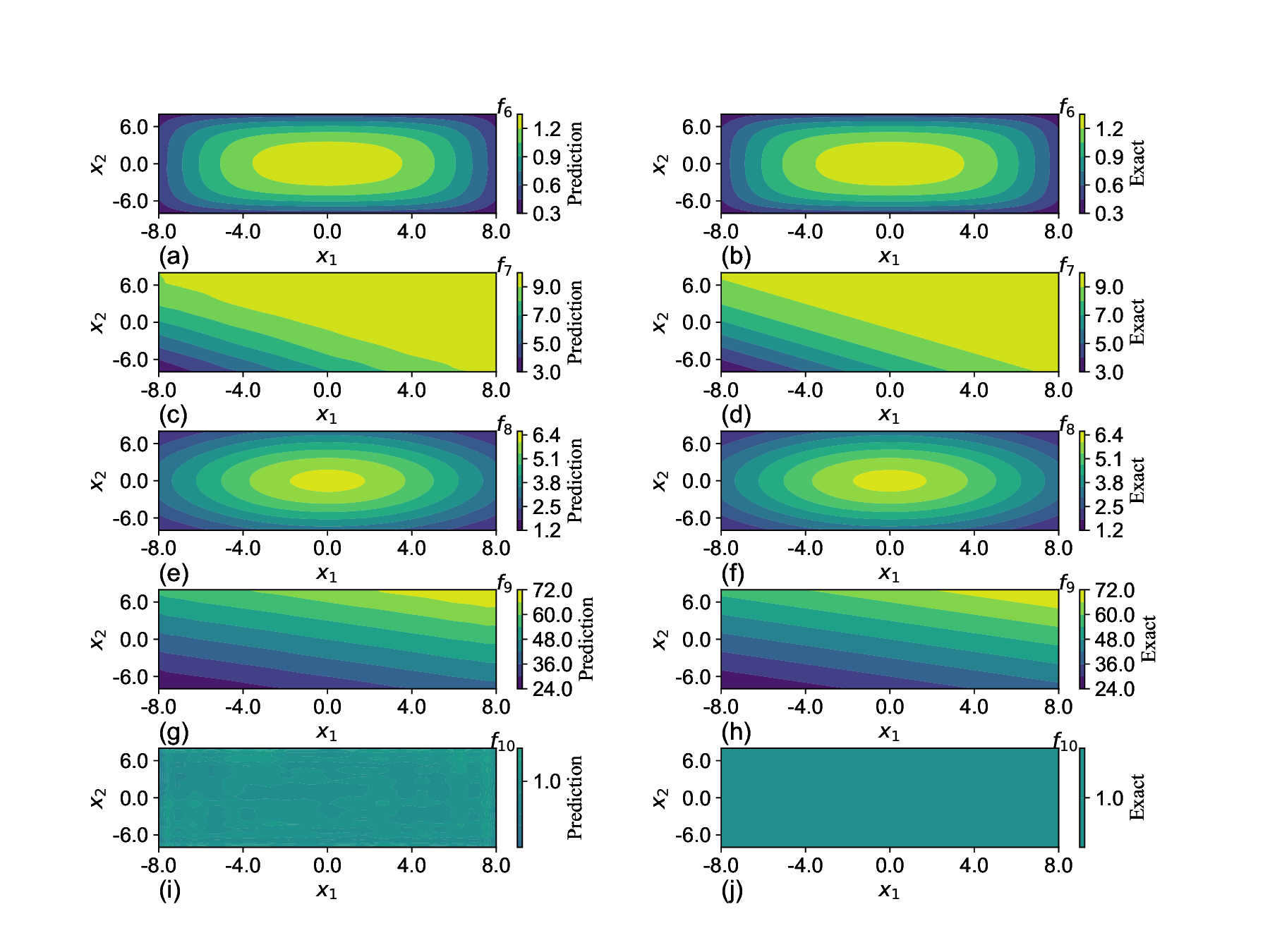}
\caption{Prediction vs. exact of $f_6$-$f_{10}$ in five dimensions. The plots are generated by projecting the surface to $x_1-x_2$ plane with other coordinates fixed to zero. Left panel: prediction; right panel: ground truth. Size of training dataset: 32768.}
\label{fig:f6tof10}
\end{figure}

 The approximation of five-dimensional functions from $f_1$ to $f_{10}$ by SGNN is illustrated through cross-sectional plots in the $x_1-x_2$ plane with three other variables fixed to zero, as shown in Figs. \ref{fig:f1tof5} and \ref{fig:f6tof10}. The left panel is for prediction, and the right panel is for ground truth. During training, uniformly-sampled training sets with the size of 32768 are separately generated for all functions in order to maintain consistency. However, fewer points can be used when the function shape is simple (e.g., sink or source). The validation set used to produce the prediction plots is generated by uniformly partitioning the subspace with a step size twice the number of neurons per layer. 

 The SGNN can accurately capture the features of all candidates regardless of their geometric complexity. Although the predictions of SGNN show minor disagreements with the ground truth when the function (e.g., $f_{10}$) is constant, the differences are less than 3\%. 

\section{Comparison of SGNN and GRBFNN}
\label{SecCompareSGNN_GRBFNN}

The performance of SGNN and GRBFNN in approximating two-dimensional and three-dimensional candidate functions are presented in Tables \ref{table:2D_candidates_performance_SGNN_vs_GRBFNN} and \ref{table:3D_candidates_performance_SGNN_vs_GRBFNN}, respectively. For comparison, the centers and width of Gaussian neurons of GRBFNN are set to be trainable variables as well. We focus on the differences of total epochs, training time per epoch, and losses for comparison. The results are obtained by averaging the results of 30 runs.

As shown in Table \ref{table:2D_candidates_performance_SGNN_vs_GRBFNN}, when approximating two-dimensional functions, SGNN can achieve comparable accuracy as GRBFNN, with differences less than one-order-of magnitude in most cases. The worse case occurs with approximating $f_1$. However, the absolute difference is around 1.0E-3, and SGNN can still give a reasonably good approximation. On the other hand, the training time per epoch of SGNN is roughly one-tenth of that of GRBFNN. 

\begin{table}[!ht]
\centering
\caption{Two-dimensional function approximations using SGNN and GRBFNN. Data is generated by averaging the results of 30 runs. Sample points: 1024; mini-batch size: 64; neurons per layer: 10.}
\label{table:2D_candidates_performance_SGNN_vs_GRBFNN}
 \vspace{5mm}
 \centering
\footnotesize{
\begin{tabular}{c|cccc|cccc}
\toprule
\multirow{2}{*}{} & \multicolumn{4}{c|}{SGNN} & \multicolumn{4}{c}{GRBFNN}\\
 & Epoch & Sec/epoch & Ave loss & Min loss & Epoch & Sec/epoch & Ave loss & Min loss\\
\hline 
$f_1$ & 483 & \textbf{0.027} & 1.86E-03 & 1.32E-03 & 568 & \textbf{0.232} & 1.98E-04 & 1.25E-04 \\
$f_2$ & 333 & \textbf{0.028} & 6.51E-05 & 3.65E-05 & 351 & \textbf{0.199} & 2.17E-05 & 1.35E-05 \\
$f_3$ & 334 & \textbf{0.027} & 1.32E-05 & 6.44E-06 & 349 & \textbf{0.193} & 6.21E-06 & 2.30E-06 \\
$f_4$ & 223 & \textbf{0.028} & 7.12E-06 & 5.53E-06 & 209 & \textbf{0.194} & 4.62E-06 & 2.68E-06 \\
$f_5$ & 123 & \textbf{0.030} & 3.72E-06 & 1.82E-06 & 126 & \textbf{0.198} & 1.99E-06 & 1.15E-06 \\
$f_6$ & 595 & \textbf{0.027} & 2.41E-04 & 1.50E-04 & 629 & \textbf{0.209} & 7.41E-05 & 3.20E-05 \\
$f_7$ & 602 & \textbf{0.027} & 6.17E-04 & 4.93E-04 & 636 & \textbf{0.192} & 1.77E-04 & 8.24E-05 \\
$f_8$ & 822 & \textbf{0.026} & 3.62E-04 & 2.70E-04 & 756 & \textbf{0.191} & 3.48E-04 & 1.56E-04 \\
$f_9$ & 625 & \textbf{0.026} & 8.56E-04 & 4.38E-04 & 591 & \textbf{0.190} & 4.84E-04 & 1.45E-04 \\
$f_{10}$ & 437 & \textbf{0.026} & 2.97E-05 & 2.26E-05 & 444 & \textbf{0.199} & 1.09E-05 & 6.15E-06\\
\bottomrule
\end{tabular}
}
\end{table}

\begin{table}[!ht]
\centering
\caption{Approximations of tri-variate functions using SGNN and GRBFNN. SGNN can achieve 100 times speedup over GRBFNN, with even smaller loss values for functions $f_3$-$f_5$ (highlighted). Data was generated by averaging the results of 30 runs. Sample points: 2048; mini-batch size: 64; Neurons per layer: 10.}
\label{table:3D_candidates_performance_SGNN_vs_GRBFNN}
 \vspace{5mm}
 \centering
\footnotesize{
\begin{tabular}{c|cccc|cccc}
\toprule
\multirow{2}{*}{} & \multicolumn{4}{c|}{SGNN} & \multicolumn{4}{c}{GRBFNN}\\
& Epoch & Sec/epoch & Ave loss & Min loss & Epoch & Sec/epoch & Ave loss & Min loss\\
\hline 
$f_1$ & 269 & \textbf{0.038} & 1.42E-03 & 9.33E-04 & 270 & \textbf{4.049} & 1.65E-04 & 9.60E-05 \\
$f_2$ & 253 & \textbf{0.039} & 2.47E-04 & 1.02E-04 & 157 & \textbf{4.055} & 4.75E-05 & 3.51E-05 \\
$f_3$ & 204 & \textbf{0.039} & \textbf{2.36E-05} & \textbf{1.78E-05} & 164 & \textbf{4.068} & 2.37E-05 & 1.80E-05 \\
$f_4$ & 188 & \textbf{0.039} & \textbf{1.82E-05} & \textbf{1.24E-05} & 97 & \textbf{4.077} & 1.96E-05 & 1.61E-05 \\
$f_5$ & 150 & \textbf{0.040} & \textbf{2.72E-06} & \textbf{1.44E-06} & 66 & \textbf{4.169} & 1.56E-05 & 1.20E-05 \\
$f_6$ & 323 & \textbf{0.037} & 7.20E-05 & 4.29E-05 & 245 & \textbf{4.085} & 3.75E-05 & 2.73E-05 \\
$f_7$ & 324 & \textbf{0.037} & 1.16E-03 & 7.07E-04 & 274 & \textbf{4.031} & 1.65E-04 & 1.02E-04 \\
$f_8$ & 315 & \textbf{0.036} & 1.95E-03 & 8.48E-04 & 314 & \textbf{3.937} & 2.03E-04 & 1.41E-04 \\
$f_9$ & 334 & \textbf{0.036} & 4.29E-03 & 1.89E-03 & 293 & \textbf{4.016} & 7.60E-04 & 4.96E-04 \\
$f_{10}$ & 227 & \textbf{0.038} & 3.28E-05 & 2.45E-05 & 188 & \textbf{4.018} & 2.63E-05 & 1.72E-05 \\
\bottomrule
\end{tabular}
}
\end{table}

The advantage of SGNN becomes more evident in three-dimensional function approximations. SGNN can gain a one-hundred-time speedup over GRBFNN but still maintain a similar level of accuracy. Surprisingly, SGNN can also yield more accurate results when approximating $f_3$ to $f_6$.
 
\section{Comparison with Deep NNs}
\label{SecCompareSGNN_DNNs}

In this section, we compare the performance of SGNN with deep ReLU and Sigmoid NNs, which are two popular choices of activation functions. Through the approximation of four-dimensional candidate functions, SGNN shows much better trainability and approximability over deep ReLU and Sigmoid NNs. 

Table \ref{table:SGNNvsReLuandSigmoid} presents the training time per epoch, total epoch for training, and loss after training of three deep NNs by averaging the results of 30 runs. All NNs possess four hidden layers with 20 neurons per layer. The training-set size is fixed to 16384, with a mini-batch size of 256. As opposed to SGNN and Sigmoid-NN, which have stable training time per epoch of across all candidate functions, the time of ReLU-NN fluctuates. This might be led by the difference in calculating derivatives of a ReLU unit with input less or greater than zero.  SGNN has a longer training time per epoch because of the computation of Gaussian function and derivative of $\mu$ and $\sigma$. One may argue that this comparison is unfair because SGNN has extra trainable variables. However, SGNN has fewer trainable weights (see Table \ref{table:SGG_vs_ReLU_configuration}) because no weights connect the input and first layer, and the output layer is not trainable.

Although SGNN has appreciably larger training epochs, this also leads to more accurate predictions. The loss values of SGNN after training are uniformly smaller than those of ReLU-NN and Sigmoid-NN except for $f_{10}$. In fact, for $f_2$, $f_4$, $f_6$, and $f_7$, the accuracy of SGNN is even two orders of magnitude better than the other two models. 

Despite the efficient training speed of Sigmoid-NN, the network is more difficult to train with random weight initialization for $f_1$ and $f_5$. In fact, the approximation of $f_5$ by Sigmoid-NN is nowhere close to the group truth after training. When functions become more complex, SGNN outperforms ReLU-NN and Sigmoid-NN in minimizing loss through stochastic gradient descent. This could be attributed to the locality of Gaussian functions that increase the active neurons, reducing the flat subspace whose gradients diminish. Sigmoid-NN aborts with significantly fewer epoch numbers. This could be led by the small derivatives of Sigmoid functions when input stays within the saturation region, which makes it more difficult to train the network.

\begin{table}[!ht]
\caption{Performance comparison of SGNN and deep neural networks with ReLU and Sigmoid activation functions. Data is generated by averaging the results of 30 runs. All NNs have four hidden layers, with 20 neurons per layer. }
\label{table:SGNNvsReLuandSigmoid}
\vspace{5mm}
\centering
\footnotesize{
\begin{tabular}{l|ccc|ccc|ccc}
\toprule
& \multicolumn{3}{c|}{SGNN} & \multicolumn{3}{c|}{ReLU-NN} & \multicolumn{3}{c}{Sigmoid-NN} \\
& Sec/epoch & Epoch & Loss & Sec/epoch & Epoch & Loss & Sec/epoch & Epoch & Loss \\ \hline
$f_1$ & 0.099 & 218 & 1.41E-03 & 0.054 & 150 & 4.86E-03 & 0.063 & 39 & \textbf{4.78E-1} \\
$f_2$ & 0.098 & 262 & 2.50E-03 & 0.054 & 119 & 1.07E-01 & 0.054 & 166 & 2.90E-01 \\
$f_3$ & 0.106 & 193 & 4.13E-05 & 0.312 & 167 & 5.22E-04 & 0.054 & 169 & 1.30E-03 \\
$f_4$ & 0.100 & 196 & 9.12E-05 & 0.234 & 161 & 7.32E-02 & 0.056 & 101 & 2.38E-01 \\
$f_5$ & 0.097 & 392 & \textbf{9.65E-02} & 0.173 & 94 & \textbf{4.97E-01} & 0.066 & 29 & \textbf{6.38E-01} \\
$f_6$ & 0.101 & 147 & 2.11E-05 & 0.293 & 115 & 1.18E-03 & 0.053 & 187 & 2.94E-03 \\
$f_7$ & 0.099 & 246 & 1.89E-03 & 0.241 & 139 & 2.07E-03 & 0.054 & 145 & 1.58E-05 \\
$f_8$ & 0.100 & 173 & 7.76E-05 & 0.344 & 109 & 9.95E-03 & 0.054 & 243 & 3.17E-03 \\
$f_9$ & 0.099 & 245 & 9.93E-03 & 0.439 & 126 & 7.66E-03 & 0.054 & 374 & 7.79E-03 \\
$f_{10}$ & 0.101 & 158 & 9.84E-06 & 0.135 & 143 & 6.86E-06 & 0.054 & 173 & 8.20E-08 \\
\bottomrule
\end{tabular}
}
\end{table}

\begin{table}[!ht]
\caption{Comparison of SGNN and ReLU-based NN in approximation of $f_{5}.$ Results are generated by averaging the data of 30 runs.}
\label{table:SGG_vs_ReLU_configuration}
\vspace{5mm}
\centering
\footnotesize{
\begin{tabular}{lccccccc}
\toprule
& Layers & Neuron/layer & Parameters & Sec/epoch & Epoch & Loss & Min loss \\
\hline
\multirow{2}{*}{SGNN} & 4 & 20 & \textbf{1360} & 0.097 & 392 & \textbf{9.65E-2} & \textbf{3.91E-2} \\
& 4 & 40 & \textbf{5120} & 0.130 & 149 & \textbf{7.08E-4} & \textbf{5.32E-4} \\
\hline
\multirow{5}{*}{ReLU-NN} & 4 & 20 & 1381 & 0.056 & 96 & 0.497 & 0.477 \\
& 4 & 40 & 5161 & 0.067 & 99 & 0.458 & 0.409 \\
& 7 & 40 & 10081 & 0.082 & 135 & 0.336 & 0.273 \\
& 10 & 40 & 15001 & 0.176 & 112 & 0.324 & 0.258 \\
& 10 & 50 & 23251 & 0.156 & 100 & 0.309 & 0.253 \\
& 10 & 60 & 33301 & 0.250 & 96 & 0.288 &0.232 \\
& 10 & 70 & 45151 & 0.120 & 97 & 0.278 &0.215 \\
& 10 & 80 &58801 & 0.261 & 89 & 0.291 &0.205\\
\bottomrule 
\end{tabular}
}
\end{table}

Next, we further compare the trainability of SGNN with ReLU-DNN. We train the two networks with different configurations to approximate the function $f_5$, which has a more complex geometry and is more difficult to approximate. The configuration of the NNs and the training performance are listed in Table \ref{table:SGG_vs_ReLU_configuration}.

Because the layer of SGNN is fixed by the number of function variables, its only tunable network hyper-parameter is the number of neurons per layer. Doubling the neurons/layer of SGNN from 20 to 40 decreases the loss by two orders of magnitude. Although the training time per epoch increases by 30\%, the number of epochs reduces by 60\%. Consequently, the total training time is cut by almost 50\%, from 38.2 to 19.4 seconds.

However, the accuracy of ReLU-DNN slightly increases with the increase of width and depth of the model. Close to 50\% loss reduction is achieved by adding 7 more layers and 50 neurons per layer. However, the error is still three orders of magnitude higher than the error by a 4-layer SGNN with one-tenth of trainable variables, with half training time per epoch.  According to the universal approximation theorem, although one can keep expanding the network structure to improve accuracy, it is against the observation in the last row. This is because the convergence of gradient descent can be a practical obstacle when the network becomes over-parametrized. In this situation, the network may impose a very high requirement on the initial weights to yield optimal solutions.

\begin{figure}
\centering
\includegraphics[width=1.0\linewidth]{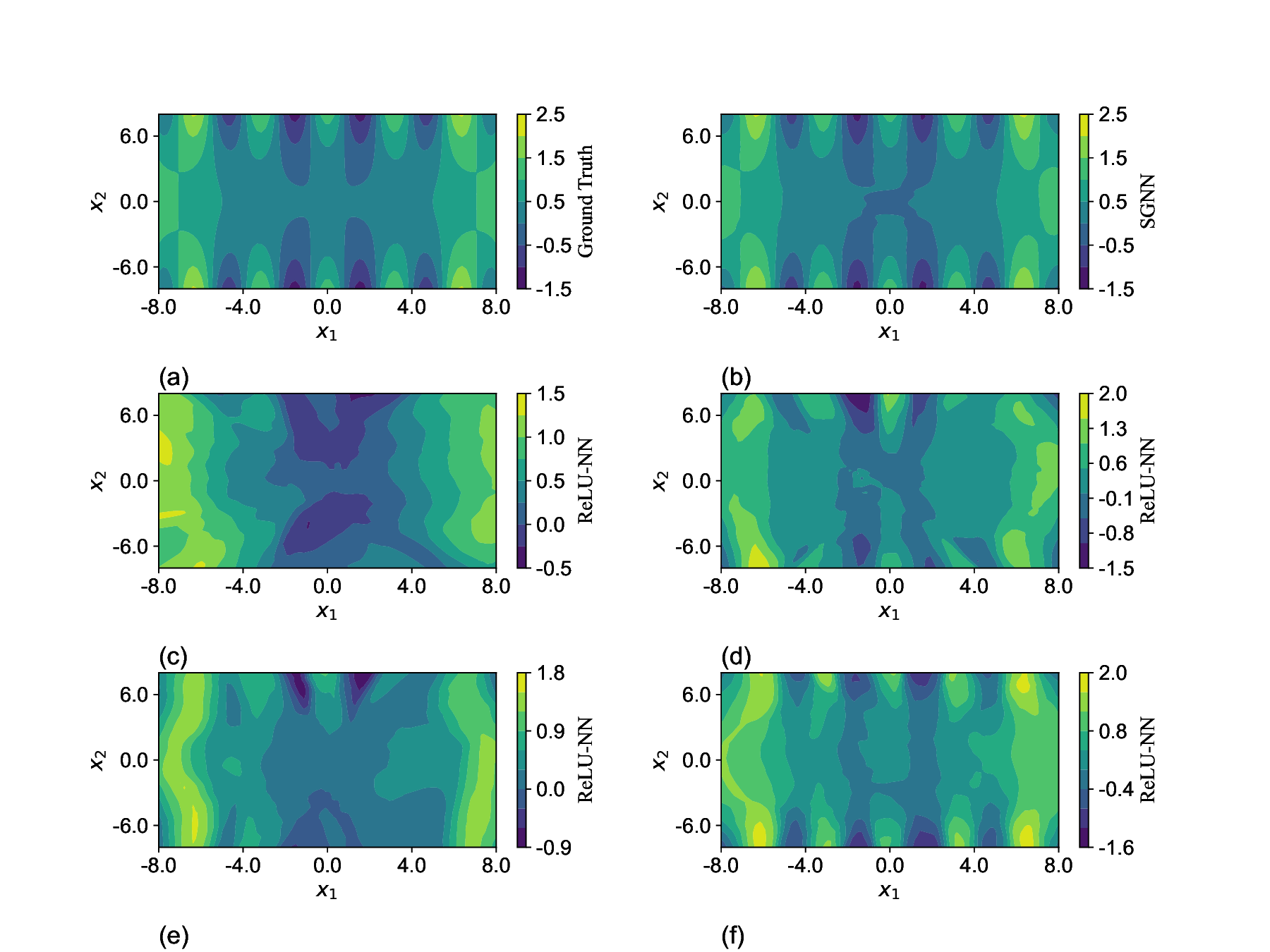}
\caption{Approximation of the four-dimensional $f_5$ using SGNN and ReLU-NNs. The plots are generated by projecting the surface to $x_1-x_2$ plane with other coordinates all zero. (a) Ground truth; (b) SGNN, (c) - (f) ReLu-NNs with different network configurations. The layers and neurons per layer of the NNs are listed in Table \ref{table:config_7}.}
\label{fig:SGNN_vs_relu_4_dim}
\end{figure}

To visualize the differences in the expressiveness between SGNN and ReLU-NN, the predictions of one run in Table \ref{table:SGG_vs_ReLU_configuration} are selected and plotted through a cross-sectional cut in $x_1-x_2$ plane with the other two variables $x_3$ and $x_4$ fixed at zero, as shown in Fig. \ref{fig:SGNN_vs_relu_4_dim}. The network configures are listed in Table \ref{table:config_7}. The predictions of SGNN in Fig. \ref{fig:SGNN_vs_relu_4_dim}(b) match well the ground truth in Fig. \ref{fig:SGNN_vs_relu_4_dim}(a). Despite the minor differences in colors near the origin, their maximum magnitude is less than 0.1. The ReLU-NN with the same structure has a much worse approximation. Although the network gradually captures the main geometric features of $f_5$ with significantly augmenting its structure to 10 layers and 70 neurons per layer, the difference of magnitude can still be as large as 0.5, as shown in Fig. \ref{fig:SGNN_vs_relu_4_dim}(f). 

\begin{table}
\centering
\caption{Network configurations of subplots of Fig. \ref{fig:SGNN_vs_relu_4_dim}.}
\label{table:config_7}
\vspace{5mm}
\begin{tabular}{cccc}
\toprule
Subfigure & Network & Layers & Neurons/Layer \\
\hline
(b) & SGNN & 4 & 40 \\
(c) & ReLU-NN & 4 & 40 \\
(d) & ReLU-NN & 7 & 40 \\
(e) & ReLU-NN & 10 & 40 \\
(f) & ReLU-NN & 10 & 70 \\
\bottomrule
\end{tabular}
\end{table}

\section{Conclusions}
\label{SecConclusion}

In this paper, we reexamined the structure of GRBFNN in order to make it tractable for problems with high-dimensional input. By using the separable property of Gaussian radial-basis functions, we proposed a new feedforward network called Separable-Gaussian-Neural-Network (SGNN), which has identical output as a GRBFNN. Different from the traditional MLPs, SGNN splits the input data into multiple columns by dimensions and feeds them into the corresponding layers in sequence. As opposed to GRBFNN, SGNN significantly reduces the number of neurons, trainable variables, and computational load of forward and backward propagation, leading to exponential improvement of training efficiency. SGNN can also preserve the dominant subspace of the Hessian matrix of GRBFNN in gradient descent and, therefore, offer comparable minimal loss. Extensive numerical experiments have been carried out, demonstrating that SGNN has superior computational performance over GRBFNN while maintaining a similar level of accuracy. In addition, SGNN is superior to MLPs with ReLU and Sigmoid units when approximating complex functions. Further investigation should focus on the universal approximability of SGNN and its applications to Physics-informed neural networks (PINNs) and reinforcement learning.

\section*{Acknowledgment}

Siyuan Xing would like to thank the Donald E. Bently center for Engineering Innovation for their kind support of teaching release time, which made this research possible.

\bibliographystyle{elsarticle-harv} 
\bibliography{cas-refs}

\end{document}